\documentclass[11pt]{wlscirep}
\usepackage[utf8]{inputenc}
\usepackage{float}
\usepackage[T1]{fontenc}
\usepackage[draft]{changes}
\usepackage{enumitem}
\usepackage{comment}
\newcommand{\citep}{\cite} %
\newcommand{\citet}{\cite} %

\usepackage{microtype}
\definecolor{citecolor}{HTML}{2779af}
\definecolor{linkcolor}{HTML}{c0392b}
\usepackage{url}
\usepackage{booktabs}

\usepackage{tabularx}
\usepackage{multirow}
\usepackage{booktabs}
\usepackage{tabulary}  %
\usepackage[abs]{overpic}
\usepackage[skins]{tcolorbox}
\usepackage{listings}
\usepackage{enumitem}
\usepackage{soul}

\definecolor{codegreen}{rgb}{0,0.6,0}
\definecolor{codegray}{rgb}{0.5,0.5,0.5}
\definecolor{codepurple}{rgb}{0.58,0,0.82}
\definecolor{backcolour}{rgb}{0.95,0.95,0.92}

\lstdefinestyle{mystyle}{
    backgroundcolor=\color{backcolour},
    commentstyle=\color{codegreen},
    keywordstyle=\color{magenta},
    stringstyle=\color{codepurple},
    basicstyle=\ttfamily\footnotesize,
    breakatwhitespace=false,         
    breaklines=true,                 
    keepspaces=true,                
    showspaces=false,                
    showstringspaces=false,
    showtabs=false,                  
    tabsize=2
}

\definecolor{lightgreen}{HTML}{EDF8FB}
\definecolor{mediumgreen}{HTML}{66C2A4}
\definecolor{darkgreen}{HTML}{006D2C}

\newcommand{\blockcomment}[1]{}

\definecolor{CMpurple}{rgb}{0.6,0.18,0.64}

\usepackage{caption}
\usepackage{verbatim}

\usepackage{dsfont}
\usepackage{amsmath}
\usepackage{amssymb}
\usepackage{mathtools}
\usepackage{amsthm}

\usepackage{listings}
\usepackage{xcolor}

\usepackage{chngcntr}
\usepackage{caption}

\DeclareCaptionType[fileext=lof,placement={!ht},within=none]{suppfigure}[Figure][List of Supplementary Figures]
\DeclareCaptionType[fileext=lot,placement={!ht},within=none]{supptable}[Table][List of Supplementary Tables]

\usepackage{hyperref}
\usepackage{nameref}

\definecolor{codegreen}{rgb}{0,0.6,0}
\definecolor{codegray}{rgb}{0.5,0.5,0.5}
\definecolor{codepurple}{rgb}{0.58,0,0.82}
\definecolor{backcolour}{rgb}{0.95,0.95,0.92}

\lstdefinestyle{mystyle}{
    backgroundcolor=\color{backcolour},
    commentstyle=\color{codegreen},
    keywordstyle=\color{magenta},
    stringstyle=\color{codepurple},
    basicstyle=\ttfamily\footnotesize,
    breakatwhitespace=false,         
    breaklines=true,                 
    keepspaces=true,                
    showspaces=false,                
    showstringspaces=false,
    showtabs=false,                  
    tabsize=2
}

\usepackage{xcolor}         %
\usepackage{tcolorbox}
\usepackage{caption}
\usepackage{wrapfig}
\usepackage{enumitem}
\usepackage{changepage}
\usepackage{subcaption}

\usepackage{colortbl} %
\definecolor{lightgray}{gray}{0.9}
\usepackage{threeparttable} %

\newtcolorbox{findingbox}{
  colback=white,
  colframe=blue!50!black,
  fonttitle=\bfseries,
  title={Finding},
  sharp corners,
  boxrule=1pt,
  left=0pt
}

\usepackage{xcolor}
\definecolor{stanfordred}{RGB}{140,21,21} %
\definecolor{weixingreen}{RGB}{0,150,0}   %
\definecolor{yuhuiblue}{RGB}{0,0,150}     %
\definecolor{binglupurple}{RGB}{128,0,128} %
\definecolor{yianorange}{RGB}{255,140,0}

\title{
\centering 
The Widespread Adoption of Large Language Model-Assisted Writing Across Society
}

\author[1*]{Weixin Liang} %
\author[2*]{Yaohui Zhang} %

\author[3]{Mihai Codreanu} %
\author[4]{Jiayu Wang} %

\author[6]{Hancheng Cao}

\author[1,2,5+]{James Zou}

\affil[1]{Department of Computer Science, Stanford University, Stanford, CA 94305, USA}

\affil[2]{Department of Electrical Engineering, Stanford University, Stanford, CA 94305, USA}

\affil[3]{Department of Economics, Stanford University, Stanford, CA 94305, USA}

\affil[4]{Paul G. Allen School of Computer Science \& Engineering, University of Washington, Seattle, WA 98195, USA}

\affil[5]{Department of Biomedical Data Science, Stanford University, Stanford, CA 94305, USA}

\affil[6]{Goizueta Business School, Emory University, Atlanta, GA 30322, USA}

\affil[*]{W.L. and Y.Z. contributed equally to this work}

\begin{abstract}

The recent advances in large language models (LLMs) attracted significant public and policymaker interest in its adoption patterns.  In this paper, we systematically analyze LLM-assisted writing across four domains—consumer complaints, corporate communications, job postings, and international organization press releases—from January 2022 to September 2024. Our dataset includes 687,241 consumer complaints, 537,413 corporate press releases, 304.3 million job postings, and 15,919 United Nations (UN) press releases.
Using a robust population-level statistical framework, we find that LLM usage surged following the release of ChatGPT in November 2022. By late 2024, roughly 18\% of financial consumer complaint text appears to be LLM-assisted, with adoption patterns spread broadly across regions and slightly higher in urban areas. For corporate press releases, up to 24\% of the text is attributable to LLMs. In job postings, LLM-assisted writing accounts for just below 10\% in small firms, and is even more common among younger firms. UN press releases also reflect this trend, with nearly 14\% of content being generated or modified by LLMs.
Although adoption climbed rapidly post-ChatGPT, growth appears to have stabilized by 2024, reflecting either saturation in LLM adoption or increasing subtlety of more advanced models. Our study shows the emergence of a new reality in which firms, consumers and even international organizations substantially rely on generative AI for communications.

\end{abstract}

\begin{document}
\maketitle
\thispagestyle{empty}

\section*{Introduction}
\label{sec:introduction}

The  emergence of large language models (LLMs) marked a significant moment in artificial intelligence, offering unprecedented capabilities in natural language processing and generation.  
This rapid proliferation of LLMs generated both excitement and concern. On one hand, LLMs have the potential to greatly enhance productivity; in the writing space specifically, it can democratize content creation (especially for non-native speakers). On the other hand, policymakers fear an erosion of trust, risks of biases and discrimination, and job displacement \cite{whitehouse2022ai,bommasani2021opportunities,StanfordAIIndex2024}; businesses worry about reliability and data privacy; academics debate the implications for research integrity and teaching \cite{Dwivedi2023,Kasneci2023}; and the public is concerned about misinformation, deepfakes, and authenticity \cite{bender2021,weidinger2021ethical}. Further complicating the discourse is the question of how LLMs may widen or potentially bridge socioeconomic gaps, given differential access to these advanced technologies.

Although some early adoption stories or isolated examples have drawn significant media attention, and survey studies have explored LLM adoption from an individual user perspective \cite{humlum2024chatgpt, bick2024rapid}, there remains a lack of systematic evidence about the patterns and extent of LLM adoption across various diverse writing domains. While some previous work used commercial software to detect such patterns \cite{brooks2024riseaigeneratedcontentwikipedia, shin2024adoption},  these studies often been constrained to single domains, relied on black-box commercial AI detectors, or analyzed relatively small datasets.
To address this gap, we conduct the first large-scale, systematic analysis of LLM adoption patterns across consumer, firm and institution communications. 
Our analysis leverages a statistical framework validated in our previous work \cite{liang2024monitoring} 
to quantify the prevalence of LLM-modified content. This framework has demonstrated superior robustness, transparency (and lower cost) compared to commercial AI content detectors \cite{liang2024monitoring, liang2024mapping, Liang2023GPTDA}, allowing us to track adoption trajectories and uncover key demographic and organizational factors driving LLM integration. 

We focus on four domains where LLMs are likely to influence communication and decision-making: consumer complaints, corporate press releases, job postings, and United Nations press releases. Consumer complaints offer insight into user–business interactions and show how these technologies may extend beyond AI-powered customer service \cite{Brynjolfsson2023}. Corporate press releases reflect strategic organizational usage, as firms incorporate LLMs into their investor relations, public relations, and broader business communications. Job postings reveal how recruiters and human resource departments harness LLMs, shedding light on broader labor market trends. Finally, UN press releases showcase the growing institutional adoption of AI for regulatory, policy, and public outreach efforts.\footnote{We also conducted a similar analysis of patent applications. However, due to the standard 18-month embargo between application and publication, our study period did not yield sufficient data to draw robust conclusions. Still, in the limited sample of late-2024 published patents, we observed a (very) moderate uptick in LLM-generated text.}

This comprehensive approach reveals several patterns. First, we observe a consistent trajectory across all the analyzed domains: rapid initial adoption following ChatGPT's release, followed by a distinctive stabilizing trend highlighting widespread adoption. One of the remarkable results from our analysis is how similar adoption is between these diverse domains. By the end of the period we analyzed, in the financial dataset we estimate about 18\% of the data was generated by LLM, around 24\% in company press releases, up to 15\% for young and small companies job postings, and 14\% for international organizations. Second, we uncover some heterogeneity in adoption rates across geographic regions, demographic groups, and organizational characteristics. Third, we find that organizational age and size emerge as the most important predictor of differential adoption, with smaller and younger firms showing markedly higher utilization rates. 

Our findings provide crucial insights into the first wave of LLM integration across society, revealing how various socioeconomic and organizational factors shape technology adoption patterns. This understanding is essential for policymakers, business leaders, and researchers as they navigate the implications of AI integration across different sectors of society and work to ensure equitable access to and responsible deployment of these powerful new tools in the future.

\section*{Results}
\label{sec:Results}

\subsection*{Widespread Adoption of Large Language Models in Writing Assistance Across Domains}

\begin{figure}[htb]
\centering
\includegraphics[width=\textwidth]{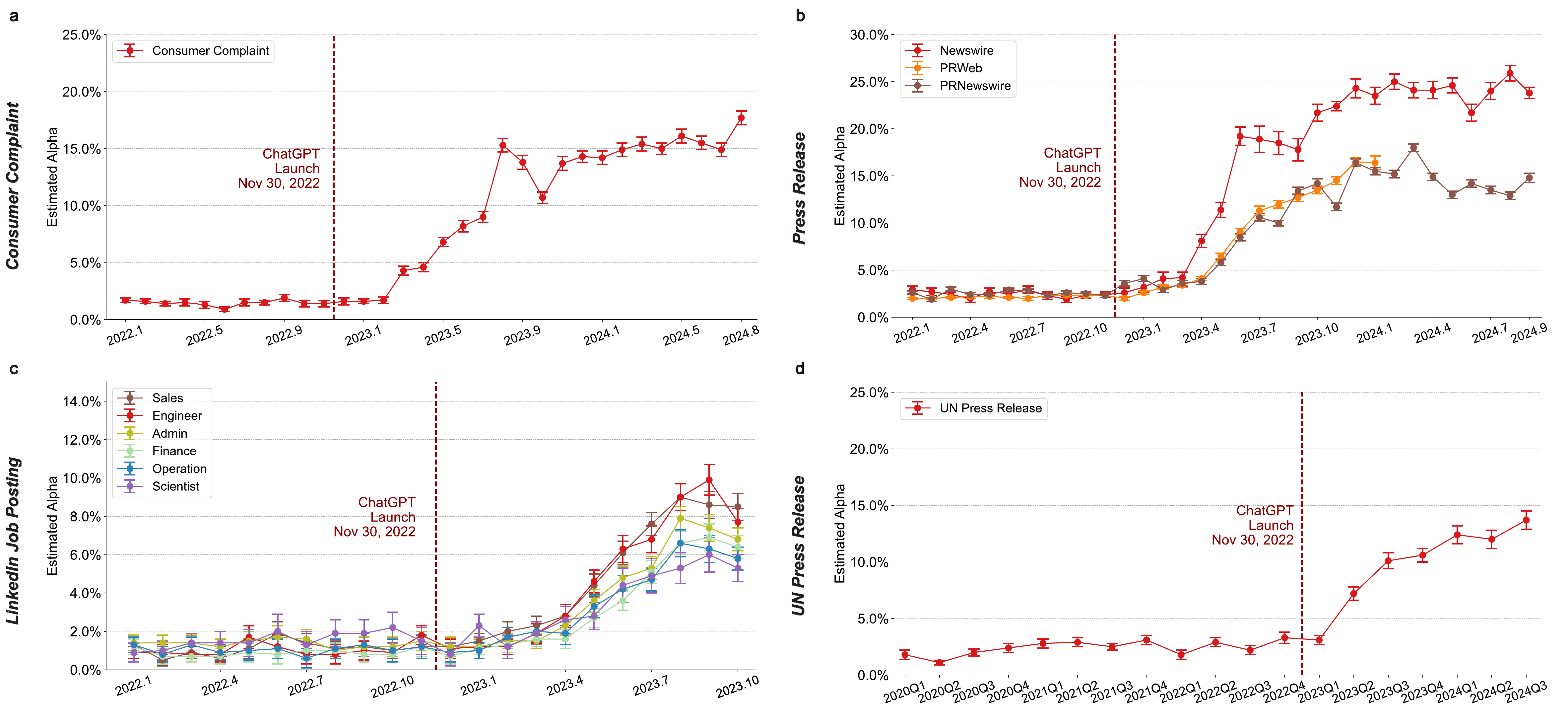}
\caption{
\textbf{Temporal dynamics of large language model (LLM) adoption across diverse writing domains.}
Analysis of LLM-generated or substantially modified content across four domains: (a) Consumer complaints filed with the Consumer Financial Protection Bureau showed algorithm false positive rate of 1.5\% pre-ChatGPT release (November 2022), followed by genuine LLM adoption rising to 15.3\% by August 2023, before plateauing at 17.7\% through August 2024. (b) Corporate press releases demonstrated consistent adoption patterns across platforms: Newswire platform showed rapid uptake reaching 24.3\% by December 2023, stabilizing at 23.8\% through September 2024; PRNewswire demonstrated similar trends with peak adoption at 16.4\% (December 2023) maintaining at 16.5\% through September 2024; PRWeb showed comparable patterns (data available through January 2024). (c) LinkedIn job postings from small organizations (below median job postings) displayed consistent trends across professional categories, with adoption increasing post-ChatGPT release (5-month lag), peaking in July 2023 before plateauing or slightly declining through October 2023. (d) United Nations government press releases showed two phases: rapid initial adoption (Q1 2023: 3.1\% to Q3 2023: 10.1\%), followed by a more gradual increase to 13.7\% by Q3 2024. This figure displays the fraction ($\alpha$) of sentences estimated to have been substantially modified by LLM using our previous method \cite{liang2024monitoring}. Error bars indicate 95\% confidence intervals by bootstrap.
}
\label{fig:main:1}
\end{figure}

We systematically analyzed large language model (LLM) adoption patterns across four distinct domains: consumer complaints, corporate PR communications, job postings, and governmental press releases (see Supplementary Information for data collection and preprocessing). Our analysis reveals a consistent pattern of initial rapid adoption following ChatGPT's release, followed by a notable stabilization period that emerged between mid to late 2023 across all domains (\textbf{Fig.~\ref{fig:main:1}}).\footnote{While the patterns across all time series show a slower adoption through 2024, these could be (at least partly) the product of more sophistication when adopting AI tools or the developments in LLMs making writing more undistinguishable from human writing.}

In the consumer complaint domain (\textbf{Fig.~\ref{fig:main:1}a}), initial LLM adoption surged about 3-4 months after the release of ChatGPT in November 2022. The proportion of content flagged as LLM-generated or substantially modified rose sharply from a baseline algorithm false positive rate of 1.5\% to 15.3\% by August 2023. This rapid growth plateaued, with only a modest increase to 17.7\% observed through August 2024.

Corporate press releases demonstrated similar adoption trends across platforms (\textbf{Fig.~\ref{fig:main:1}b}), once again about 3-4 months post-ChatGPT release. Newswire saw rapid growth, peaking at 24.3\% in December 2023 and stabilizing at 23.8\% through September 2024. PRNewswire followed closely, reaching 16.4\% in December 2023 and maintaining this level through September 2024. PRWeb exhibited comparable dynamics, with data available through January 2024.

LinkedIn job postings from small organizations showed profession-specific adoption trends but similarly reflected a slowing trajectory (\textbf{Fig.~\ref{fig:main:1}c}). Following a five-month lag post-ChatGPT release, adoption increased steadily across professional categories, peaking in July 2023 between 6-10\%. These figures are higher in the sample of small and young firms, where they reach more than 10\%, and up to 15\% (\textbf{Fig.~\ref{fig:main:4}})
. Adoption rates either plateaued or showed signs of slight declines through October 2023, when the latest data was available. 

International organization communication, here measured by United Nations press releases by country teams followed a similar adoption pattern with initial rapid growth that later plateaued (\textbf{Fig.~\ref{fig:main:1}d}). The initial phase was marked by a rapid increase from 3.1\% in Q1 2023 to 10.1\% in Q3 2023. This was followed by a slower, incremental rise, reaching 13.7\% by Q3 2024. 

\subsection*{Geographic and Demographic Disparities in Consumer Complaint LLM Adoption}

\begin{figure}[htb]
\centering
\includegraphics[width=\textwidth]{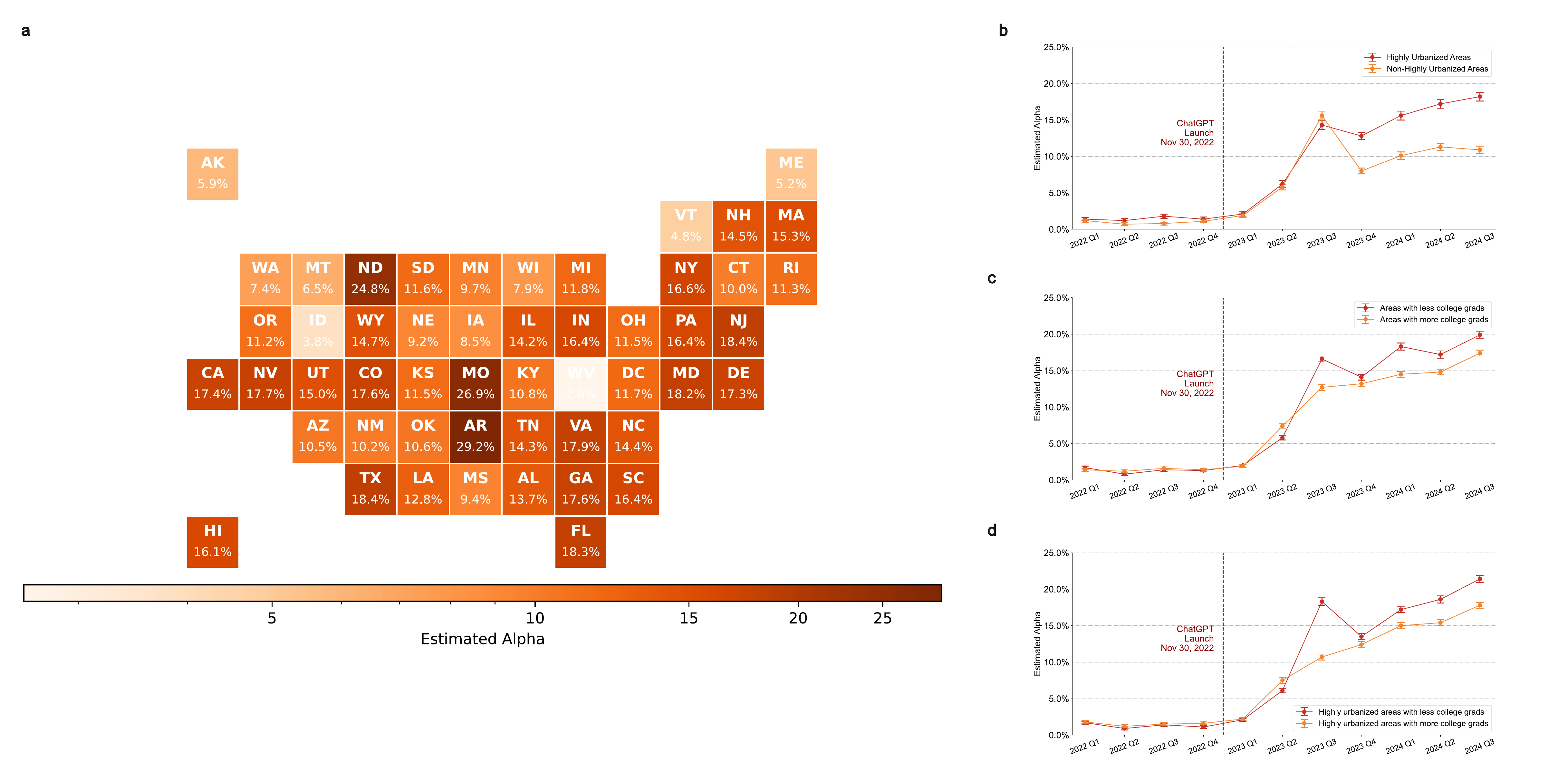}
\caption{
\textbf{Geographic and demographic patterns of LLM adoption in Consumer Financial Protection Bureau complaints.}
(a) State-level analysis (January-August 2024) revealed substantial geographic variation, with highest adoption in Arkansas (29.2\%), Missouri (26.9\%), and North Dakota (24.8\%), contrasting with lowest rates in West Virginia (2.6\%), Idaho (3.8\%), and Vermont (4.8\%). Notable population centers showed moderate adoption (California: 17.4\%, New York: 16.6\%). (b) Analysis by Rural Urban Commuting Area (RUCA) codes showed similar adoption trajectories between highly urbanized and non-highly urbanized areas during initial uptake (2023Q1-2023Q3), before diverging to equilibrium levels of 18.2\% and 10.9\%, respectively. (c) Comparison of areas above and below state median levels of bachelor's degree attainment (population aged 25+) revealed comparable initial adoption patterns (2023Q1-2023Q2), followed by higher stabilized rates in areas with lower educational attainment (19.9\% vs 17.4\% by 2024Q3). (d) Within highly urbanized areas, this educational attainment pattern persisted, with lower-education areas showing higher adoption rates (21.4\% vs 17.8\% by 2024Q3). 
}
\label{fig:main:2}
\end{figure}

Our analysis of Consumer Financial Protection Bureau complaints revealed some geographic and demographic heterogeneity in LLM adoption patterns (\textbf{Fig.~\ref{fig:main:2}}).  At the state level, we observed variation in adoption rates during the January-August 2024 period, with highest adoption in Arkansas (29.2\%, 7,376 complaints), Missouri (26.9\%, 16,807 complaints), and North Dakota (24.8\%, 1,025 complaints). This contrasted sharply with minimal adoption in West Virginia (2.6\%, 2,010 complaints), Idaho (3.8\%, 1,651 complaints), and Vermont (4.8\%, 361 complaints). Notably, major population centers demonstrated much less variation in adoption levels, with California (157,056 complaints) and New York (104,862 complaints) showing rates of 17.4\% and 16.6\%, respectively (\textbf{Fig.~\ref{fig:main:2}a}). However, this could be interpreted either as a genuine differential compared to the smaller states in the left and right tail or the product of lower sample noise (due to higher number of observations).

The adoption of LLMs varied over time between more and less urbanized areas. Analysis using Rural Urban Commuting Area (RUCA) codes showed that highly urbanized and non-highly urbanized areas initially displayed similar adoption trajectories during the early phase (2023Q1-2023Q3). However, these trajectories subsequently diverged, reaching equilibrium levels of 18.2\% in highly urbanized areas compared to 10.9\% in non-highly urbanized areas (\textbf{Fig.~\ref{fig:main:2}b}). These differences were highly statistically significant at all conventional levels.

Areas with lower educational attainment showed somewhat higher LLM adoption rates in consumer complaints. Comparing areas above and below state median levels of bachelor's degree attainment, areas with lower educational attainment ultimately stabilized at rates of around 19.9\% in 2024Q3 (compared with 17.4\%) (\textbf{Fig.\ref{fig:main:2}c}). This pattern persisted even within highly urbanized areas, where lower-education regions demonstrated higher adoption rates (21.4\% versus 17.8\% by 2024Q3) (\textbf{Fig.\ref{fig:main:2}d}). In both comparison, the p-values were less than 0.001, indicating statistically significant differences, despite qualitatively similar trends.

\subsection*{LLM Adoption in Corporate Press Releases}
\begin{figure}[htb]
\centering
\includegraphics[width=\textwidth]{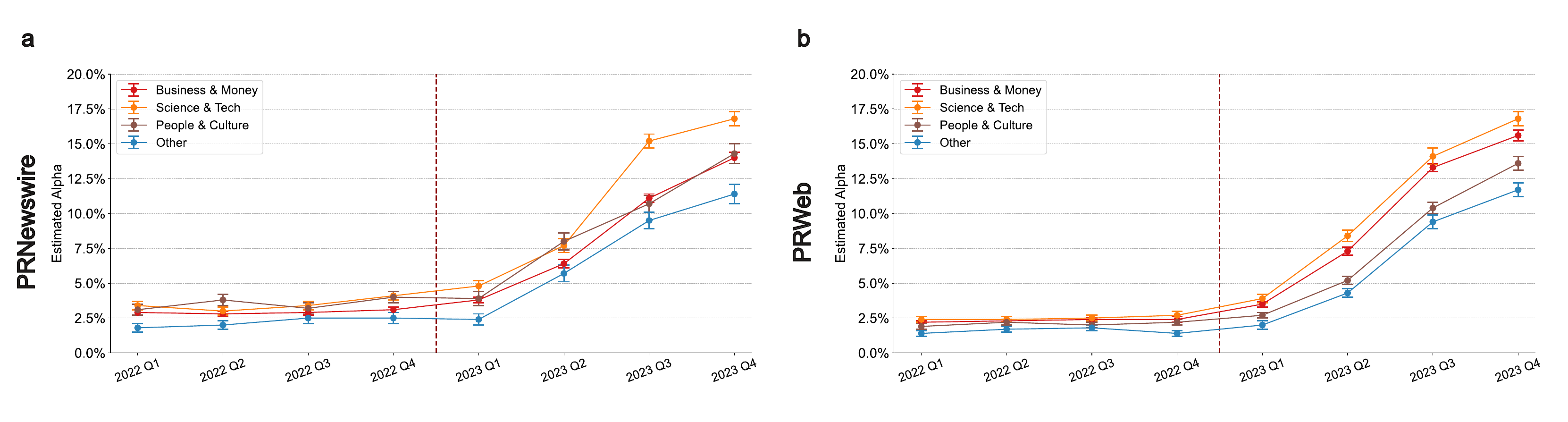}
\caption{
\textbf{Sectoral patterns of LLM adoption in corporate press releases across major distribution platforms.}
Analysis of press releases by sector revealed consistent patterns across platforms, with Science \& Technology showing marginally higher adoption rates. (a) PRNewswire demonstrated similar sectoral patterns by 2023Q4: Science \& Technology (16.8\%), People \& Culture (14.3\%), Business \& Money (14.0\%), and Other sectors (11.4\%). (b) PRWeb exhibited comparable sectoral distribution: Science \& Technology (16.8\%), Business \& Money (15.6\%), People \& Culture (13.6\%), and Other sectors (11.7\%). All sectors showed similar temporal adoption patterns following ChatGPT's release, with initial lag followed by sustained growth through 2023. 
}
\label{fig:main:3}
\end{figure}

After characterizing consumer-side adoption patterns, we next examined corporate LLM usage across major corporate press release platforms---Newswire, PRWeb, and PRNewswire, each of which caters to different audiences and industries (\textbf{Fig.~\ref{fig:main:1}b}, \textbf{Fig.~\ref{fig:main:3}a-b}).\footnote{A vast oversimplification based on available data would be that PRNewswire generally targets larger corporations with extensive reach to major news outlets and traditional media. PRWeb offers a more affordable, online-focused option with an emphasis on SEO, catering to smaller businesses. Newswire reaches both traditional and online platforms. All three offer some editorial services but focus primarily on distribution of the contents produced by the businesses.}

Before the launch of ChatGPT, the fraction of AI-modified content remained consistently low across all these sources, fluctuating around the 2-3\% mark (i.e., false positives). However, following the launch, a significant increase in AI-modified content was observed across all domains, about 2 quarters post rollout. Newswire, in particular, experienced the most dramatic rise, with the estimated fraction peaking at over 25\% by late-2023. PRWeb and PRNewswire also saw notable growth, though to a lesser degree, plateauing around 15\%. This suggests a widespread uptake of LLM technology in content creation across different types of press releases starting in early 2023.

In \textbf{Fig.~\ref{fig:main:3}a-b}, we show the quarterly growth of LLM usage in press releases across different categories for PRNewswire (a) and PRWeb (b). Both charts show a sharp rise in AI-modified content starting in early 2023, with some differential patterns emerging by topic. In both platforms, the categories "Business \& Money" and "Science \& Tech" exhibit the most pronounced increase, with Science \& Tech reaching just below 17\% by Q4 2023. "People \& Culture" and "Other" categories also demonstrate growth, but at a somewhat slower pace, which may be indicative that LLM adoption has been particularly strong in more technical and business-focused content.

Overall, we show a significant uptick in LLM writing across various press release categories. On one hand, the sharp increase in AI-modified content in press releases suggests that businesses are leveraging LLMs to improve efficiency in content creation. By utilizing AI, companies can potentially produce high-quality communications more quickly and cost-effectively, especially in areas requiring frequent updates and complex information dissemination. This may also be advantageous if companies are trying to withhold more sensitive information from the public and use more generic language. On the other hand, the growing reliance on AI-generated content may introduce challenges in communication. In sensitive categories, over-reliance on AI could result in messages that fail to address concerns or overall release less credible information externally. Over-reliance on AI could also introduce public mistrust in the authenticity of messages sent by firms.

\subsection*{LLM Adoption in LinkedIn Job Postings}

We next examined another dimension of corporate LLM adoption through analysis of LinkedIn job postings.
We first took the whole sample of LinkedIn job posting and analyzed the effects (\textbf{Supp. Fig.~\ref{fig: full-sample-LinkedIn}}). In this full sample, we see that about 3-4\% of all vacancy postings have LLM modified content. Albeit a small increase, this is generally statistically different from pre-ChatGPT introduction (i.e. false positive) levels (with p-values less than 0.001 across categories).  However, this broader sample heavily features larger firms that post more vacancies and have greater financial and human resources to customize those postings. Such firms may also advertise the same position multiple times throughout the year and rely on their established reputation, reducing the need to update job postings frequently. Consequently, for the remainder of this analysis, we focused on small companies, defined either as firms which post less than the median number of vacancies (2 or less each year), or as businesses with 10 or fewer registered employees in 2021 or those posting two or fewer positions per year on LinkedIn (see Supplementary Information).

\begin{figure}[htb]
\centering
\includegraphics[width=\textwidth]{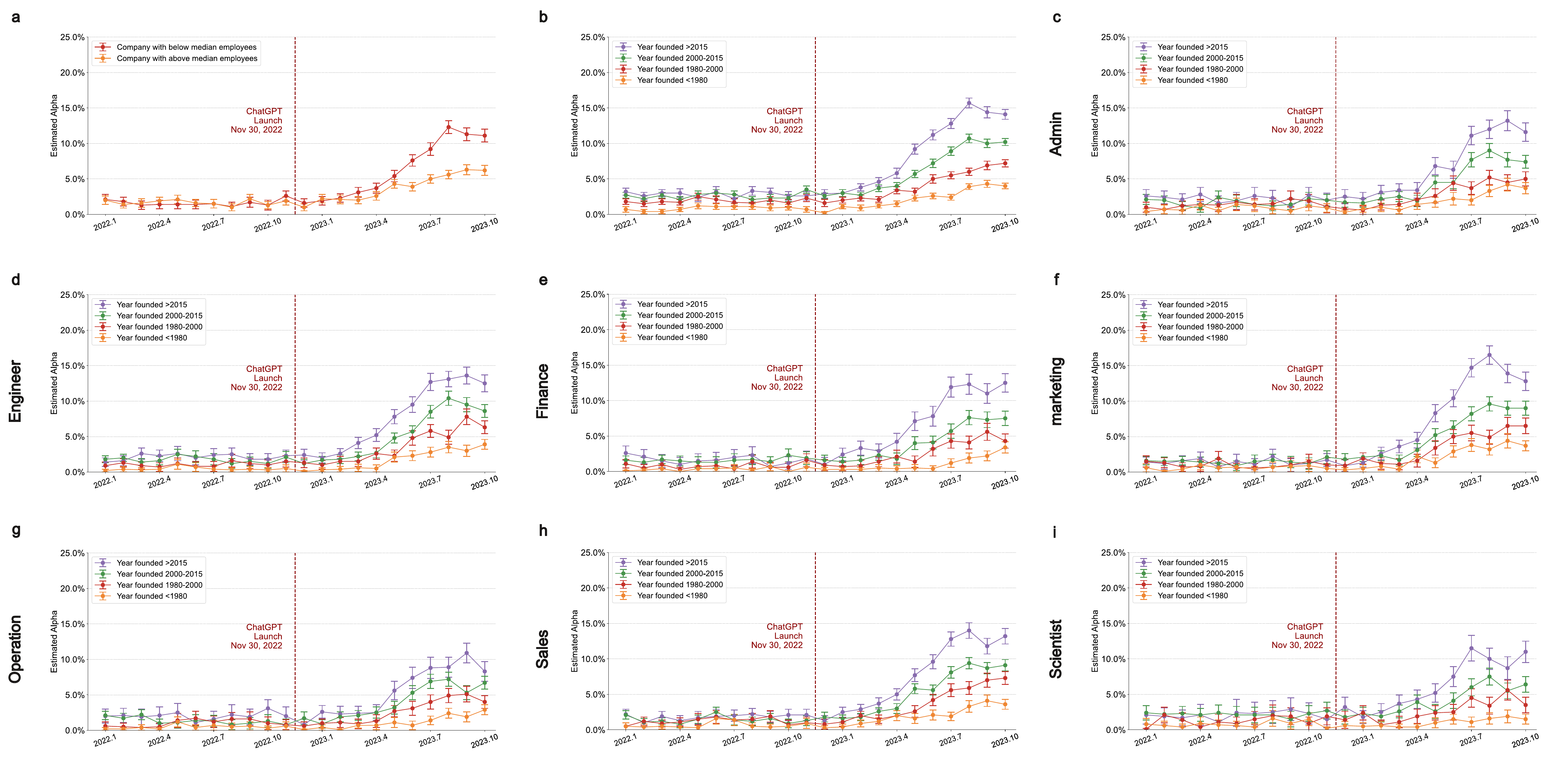}
\caption{
\textbf{Organization age and LLM adoption patterns in LinkedIn job postings from small organizations across professional categories.}
(a) Among small organizations (less than median job vacancies), analysis stratified by number of employees revealed higher LLM adoption rates in firms with below median employees (11.1\% vs 6.2\% by October 2023). (b) Among small organizations (less than median job vacancies), analysis stratified by founding year revealed higher LLM adoption rates in more recently established firms (founded after 2015: 14.1\%; 2010-2015: 10.2\%; 1980-2000: 7.2\%; pre-1980: 4.0\%). (c-i) This age-dependent pattern persisted across professional categories: Admin (c), Engineer (d), Finance (e), Marketing (f), Operations (g), Sales (h), and Scientist (i), with newer organizations consistently showing higher adoption rates. We defined small organizations based on having 2 or less job vacancy postings in a year (median is 3).
}
\label{fig:main:4}
\end{figure}

Using the sample of small companies based on the number of vacancies posted, our findings reveal a gradual but notable increase in the estimated fraction of AI-modified content for several job categories (\textbf{Fig.~\ref{fig:main:1}d}, \textbf{Fig. \ref{fig:main:4}}). Prior to the launch of ChatGPT, the fraction of AI-modified text hovered between 0–2\% across all categories, reflecting the range of false positives. After ChatGPT became available, a discernible uptick begins around early to mid-2023, leveling off by October 2023 at roughly 5–10\% for all categories. The increase is most pronounced in engineering and sales postings, which each approach 10\% AI-modified content. Finance, Admin, Scientist, and Operations show a somewhat slower growth, albeit the differences between these categories are small. If instead we define small companies by the number of employees (\textbf{Fig.~\ref{fig: supp-robust-small-company-definition}}) the Scientist category ranks first. \footnote{This may be some evidence that firms requiring more advanced scientific, financial, or marketing expertise might be more inclined to adopt AI technologies, although the differences are modest.}

We further stratified these small firms by founding year—grouping them into post‐2015, 2000–2015, 1980–2000, and pre‐1980 cohorts (\textbf{Fig.\ \ref{fig:main:4}}), based on the rough quartiles in the data. Across every job category, more recently founded companies consistently exhibit both the highest levels and the fastest uptake of LLM‐related text generation, especially following ChatGPT’s launch. Firms founded after 2015 reach 10–15\% AI‐modified text in certain roles, whereas those founded between 2000 and 2015 show moderate growth of 5–10\%. By comparison, firms founded before 1980 typically remain below 5\%. These results underscore how younger firms—possibly with younger workforces—more readily integrate new AI technologies into their hiring and onboarding processes, whereas older organizations may adopt such tools more conservatively. Overall, firm age and size emerge as (perhaps the most) significant correlates of the heterogeneity observed in LLM uptake throughout our analyses.

This trend highlights a potential shift in recruitment practices among small firms, showcasing a growing reliance on AI-writing tools. On one hand, this can decrease company hiring costs, with smaller and younger enterprises being more likely to leverage advanced tools to remain competitive despite perhaps being more liquidity constrained. On the other hand, the adoption of LLM writing in job posting could either enhance or decrease the efficiency and effectiveness in attracting qualified candidates. For jobseekers, one possible negative effect is harder differentiation between posting firms quality and position requirements.

The leveling off or even slight decrease in AI-modified content by October 2023 might indicate that the adoption rate has stabilized, potentially reaching a saturation point where firms comfortable with AI have already adopted it. Alternatively, this can be explained by increased sophistication and subtlety of these methods. Overall, the increased integration of AI in job postings suggests a transformative period in hiring, with AI playing an important role in how small firms communicate job opportunities. This could have implications for job seekers as well, who may encounter more uniformly crafted postings and might need to adapt their application strategies accordingly.

\subsection*{LLM Adoption in United Nations Press Releases}

United Nations press releases exhibited a similar two-phase adoption pattern, with an initial surge from 3.1\% to 10.1\% in Q1-Q3 2023, followed by a more gradual increase to 13.7\% by Q3 2024 (\textbf{Fig.~\ref{fig:main:1}d}). 
Across UN Member States country teams, we observed consistent adoption patterns across regions, with adoption rates reaching 11-14\% by 2024, with the exception of the UN teams in Latin American and Caribbean countries that had slightly higher adoption rates at about 20\% (\textbf{Supp. Fig.~\ref{fig: supp-robust-US-country-groups}}). The steady growth across regions reflects how LLMs are being integrated globally, even in contexts of sensitive, high-stakes communication. 

This rapid uptake suggests that country teams have found LLMs valuable for producing timely updates, which can be especially useful during pressing crises. On the other, this trend raises questions about how LLMs might affect the authenticity of vital international communication. As the UN continues to refine its stance on AI, this highlights a broader trend: even the world’s most prominent international bodies are using LLMs in their communications--underscoring both the perhaps inevitability of AI-driven writing and the questions it raises about authenticity and accountability.

\section*{Discussion}

Our findings reveal widespread adoption of large language models across diverse writing domains, ranging consumers, firms and international organizations. This finding complements and extends our previous research that found widespread adoption across academic researchers.\cite{liang2024monitoring} 
A consistent temporal pattern emerges from our data: after an initial lag of 3–4 months following the ChatGPT launch, there was a sharp surge in LLM usage, which then stabilized by late 2023 and remained steady through 2024. This trajectory deviates from traditional diffusion models that predict continuous and gradual growth, suggesting several possibilities. Early adopters may have already reached a saturation point within their domains, or domain-specific barriers (generally, these can range from costs of adoption, regulatory constraints, concerns over authenticity coupled with advances in users recognizing AI writing, etc.) that could be impeding further expansion. Alternatively, improvements in LLM sophistication may be rendering AI-generated content increasingly indistinguishable from human writing, complicating our ability to measure ongoing adoption.

In the consumer complaint domain, the geographic and demographic patterns in LLM adoption present an intriguing departure from historical technology diffusion trends \cite{rogers2014diffusion, bloom2021diffusion, 10.1093/qje/qjaf002} and technology acceptance model \cite{davis1989technology, venkatesh2003user}, where technology adoption has generally been concentrated in urban areas, among higher-income groups, and populations with higher levels of educational attainment \cite{rojas2017demographics, foster2010microeconomics}. While the urban-rural digital divide seems to persist, our finding that areas with lower educational attainment showed modestly higher LLM adoption rates in consumer complaints suggests these tools may serve as equalizing tools in consumer advocacy. This finding aligns with  survey evidence indicating  that younger, less experienced workers may be more likely to use ChatGPT \cite{humlum2024chatgpt}. %
This democratization of access underscores the potentially transformative role LLMs could play in amplifying underserved voices. However, further study is needed to assess whether this increased adoption translates into more effective consumer outcomes.

Corporate communication channels also demonstrated widespread but decelerating LLM integration. The plateauing adoption across platforms like Newswire, PRWeb, and PRNewswire raises important considerations about the balance between cost efficiency and authenticity. While LLMs may enable rapid, cost-effective content generation, overreliance on automated tools could compromise the nuance and credibility required in professional communications, potentially eroding trustworthiness \cite{jakesch2019ai, hong2018bias, kadoma2024generative}. Future research should explore how organizations navigate this trade-off and whether editorial interventions are employed to mitigate potential drawbacks.

In the recruitment process, small firms, particularly those founded after 2015, exhibited the fastest adoption of LLM-generated content. This trend suggests that younger, or companies closer to the technological frontier are leveraging LLMs to streamline hiring processes and reduce costs.\footnote{This is consistent with previous research finding that younger firms may invest relatively more in AI, but may superficially seem in contrast with datapoints showing larger firms and firms with higher cash holdings increase their AI investments more.\cite{babina2024} We think that it is possible that younger, smaller firms may use more cost-effective AI products such as ChatGPT, and may also have a lower time from AI usage to output.} While our study did not directly measure homogenization, prior research on the homogenization of LLM-generated content in academia~\cite{liang2024monitoring,liang2024mapping} suggests that similar effects could occur in job postings. This potential homogenization may inadvertently obscure critical distinctions between roles and organizations, potentially complicating job seekers' decision-making. In fact, recent evidence has shown that while employers who leverage LLM to generate first draft of job post may receive more applications, they are less likely to make a hire\cite{wiles2023impact}. Further investigating how AI-generated postings influence applicant perceptions and hiring could provide valuable insights into the long-term implications of this shift.

International institutions communication, exemplified by United Nations press releases, also demonstrated significant LLM adoption. These patterns remained robust when stratifying by regional country groups (\textbf{Supp. Fig.\ref{fig: supp-robust-US-country-groups}}). 
The presence of LLM-generated content within such formal and traditionally cautious institutions suggests that AI-driven tools are gradually influencing even high-stakes communication channels, reflecting the broad and expanding reach of these technologies. As it was the case with corporate communications, these findings raise the same trade-off between cost-efficiency and credibility.

The stabilization of LLM adoption may reflect either the maturation of AI integration or domain-specific friction factors. As LLM technologies continue to evolve, future research should aim to disentangle the drivers of adoption plateaus by examining whether they stem from market saturation, improvements in LLM indistinguishability, or external barriers. They should evaluate the impact of LLM-generated content on communication quality, credibility, and user engagement across sectors and investigate potential homogenization effects in job postings and other domains to assess how uniform AI-generated content might affect decision-making and market dynamics.

Our study has several limitations. While we focused on widely used LLMs like ChatGPT, which account for a significant portion of global usage~\cite{vanrossum2024generative}, we acknowledge that other models also contribute to content generation across domains. Additionally, although prior research has shown that GPT-detection methods can sometimes misclassify non-native writing as AI-generated~\cite{Liang2023GPTDA}, our findings consistently indicated low false positive rates during earlier periods. However, shifts in user demographics or language usage could still influence detection accuracy~\cite{Globalaitalent}. 

Perhaps the biggest limitation in our study is that we cannot reliably detect language that was generated by LLMs, but was either heavily edited by humans or was generated by models that imitate very well human writing. Therefore, one way to interpret our study is as a lower bound of adoption patterns.
Finally, our analysis primarily focuses on English-language content, potentially overlooking adoption trends in non-English-speaking regions. Future research could expand on these findings by incorporating multilingual data and refining detection methodologies.

In conclusion, we show that LLM writing is a new pervasive reality across consumer, corporate, recruitment, and even governmental communications. As these technologies continue to mature, understanding their effects on content quality, creativity, and information credibility will be critical. Addressing the regulatory and ethical challenges associated with AI-generated content will also be essential for ensuring that the benefits of LLMs are realized while maintaining transparency, diversity, and public trust in communication.

\subsection*{Acknowledgments}
We thank Daniel McFarland, Dan Jurafsky, Yian Yin, Mirac Suzgun , Zachary Izzo for their helpful comments and discussions. We thank Todd Hines and the Stanford GSB Library for help with obtaining data. J.Z. is supported by the National Science Foundation (CCF 1763191 and CAREER 1942926), the US National Institutes of Health (P30AG059307 and U01MH098953) and grants from the Silicon Valley Foundation and the Chan-Zuckerberg Initiative.

\section*{Data Availability}

The datasets analyzed in this study are publicly or privately accessible through the following sources: the \textit{Consumer Complaint Data} is available at the Consumer Financial Protection Bureau’s website (linked \href{https://www.consumerfinance.gov/data-research/consumer-complaints/}{\texttt{{here}}}); \textit{Corporate Press Release} data can be accessed via the News API (linked \href{https://www.newsapi.ai/}{\texttt{here}}); and \textit{LinkedIn Job Posting} data is obtained from the Revelio Labs Workforce Data, licensed through Stanford GSB Library (information \href{https://www.data-dictionary.reveliolabs.com/}{\texttt{here}}). Data for \textit{UN Press Releases} was directly scraped from official UN websites (e.g., \href{https://china.un.org/en/press-centre/press-releases}{\texttt{{here}}}).

\section*{Code Availability}

The code can be accessed at this GitHub \href{https://github.com/Weixin-Liang/LLM-widespread-adoption-impact}{\texttt{link}.}

\clearpage

\bibliography{main_journal}
\clearpage

\clearpage
\newpage

\clearpage
\newpage

\clearpage
\newpage

\clearpage
\newpage

\clearpage

\begin{suppfigure}[htb!]
\centering
\includegraphics[width=1.00\textwidth]{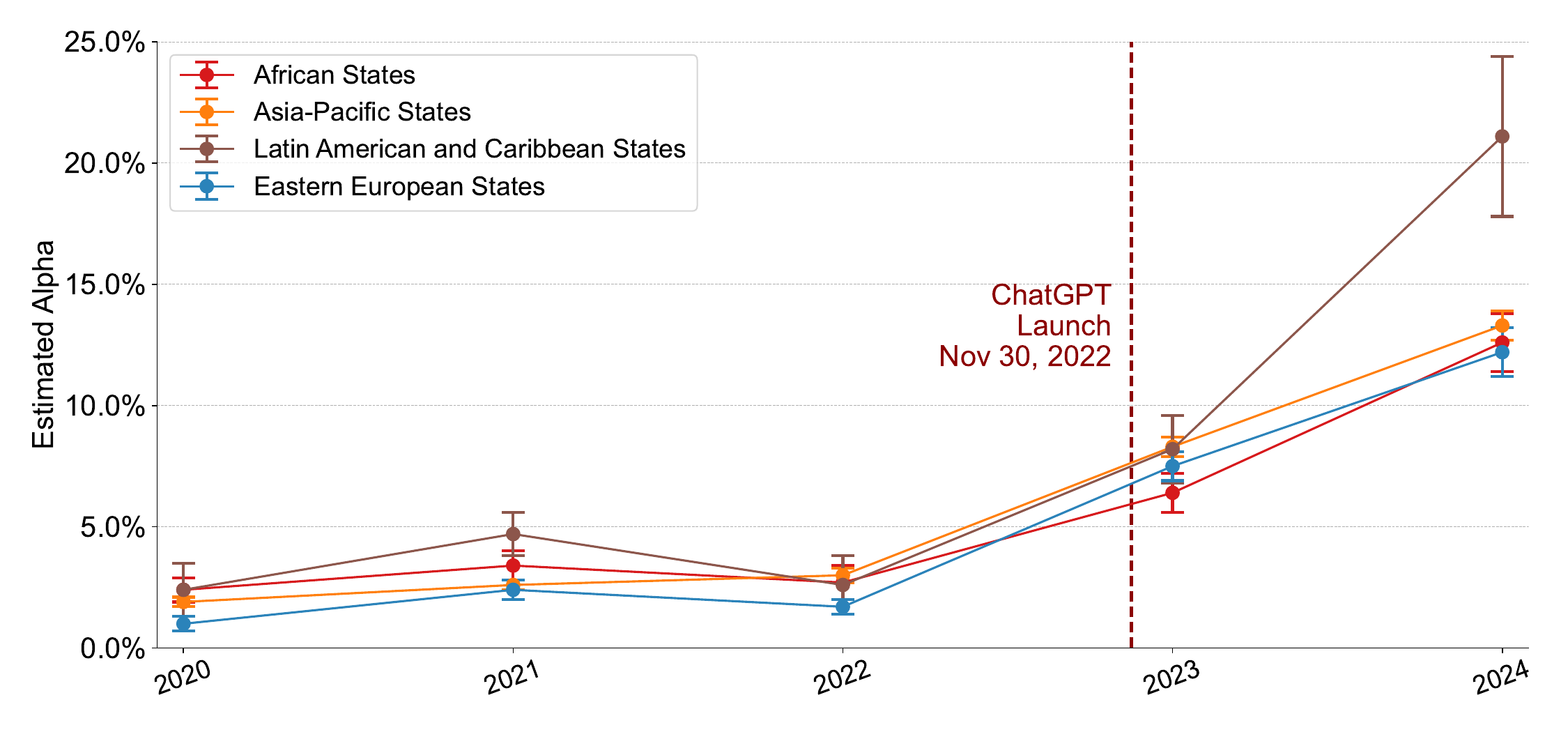}
\caption{
\textbf{Regional variation in LLM adoption across United Nations Member States' press releases.} 
Temporal analysis of estimated fraction ($\alpha$) of LLM-modified content stratified by regional groups shows differential adoption patterns. After ChatGPT's launch (November 30, 2022), Latin American and Caribbean States demonstrated the highest adoption rate, reaching approximately 21\% by 2024, while African States, Asia-Pacific States, and Eastern European States showed more moderate increases to 11-14\%. Error bars indicate 95\% confidence intervals obtained through bootstrap analysis. Regional variations may reflect differences in technological infrastructure, language diversity, and institutional policies across Member States.
}
\label{fig: supp-robust-US-country-groups}
\end{suppfigure}

\clearpage

\begin{suppfigure}[ht!]
    \centering
    \includegraphics[width=0.5\textwidth]{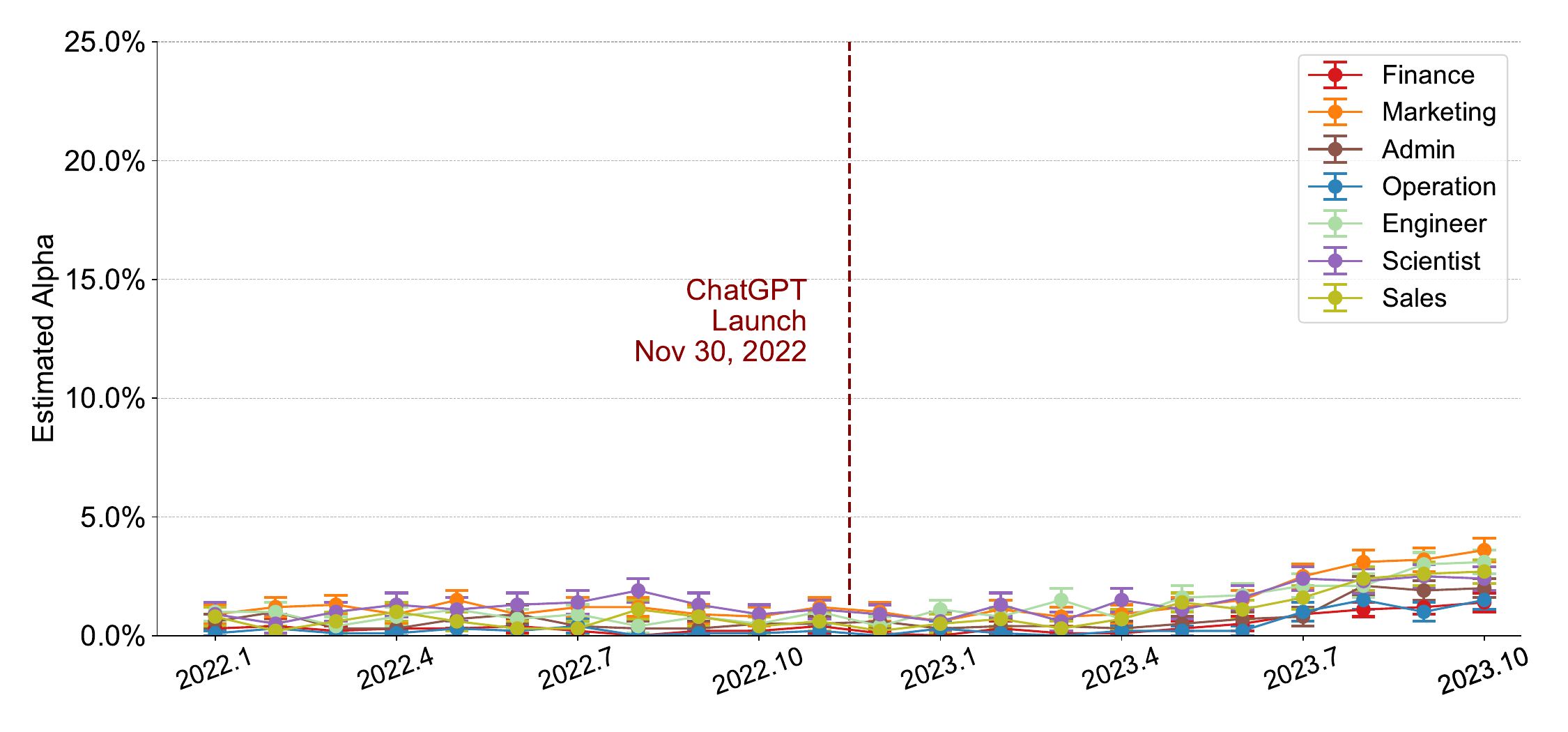}
\caption{
\textbf{Analysis of LLM adoption in LinkedIn job postings across the full sample.} Temporal analysis of the estimated fraction ($\alpha$) of LLM-modified content in job postings across all company sizes shows a modest but statistically significant increase from pre-ChatGPT baseline to approximately 3\% adoption following ChatGPT's introduction (November 30, 2022). This aggregate analysis includes all companies regardless of size, with larger firms (who post more frequent vacancies and typically have dedicated HR resources) representing a greater proportion of the sample. Error bars represent 95\% confidence intervals obtained through bootstrap analysis.
}
\label{fig: full-sample-LinkedIn}
\end{suppfigure}

\begin{suppfigure}[ht!]
    \centering
    \includegraphics[width=0.5\textwidth]{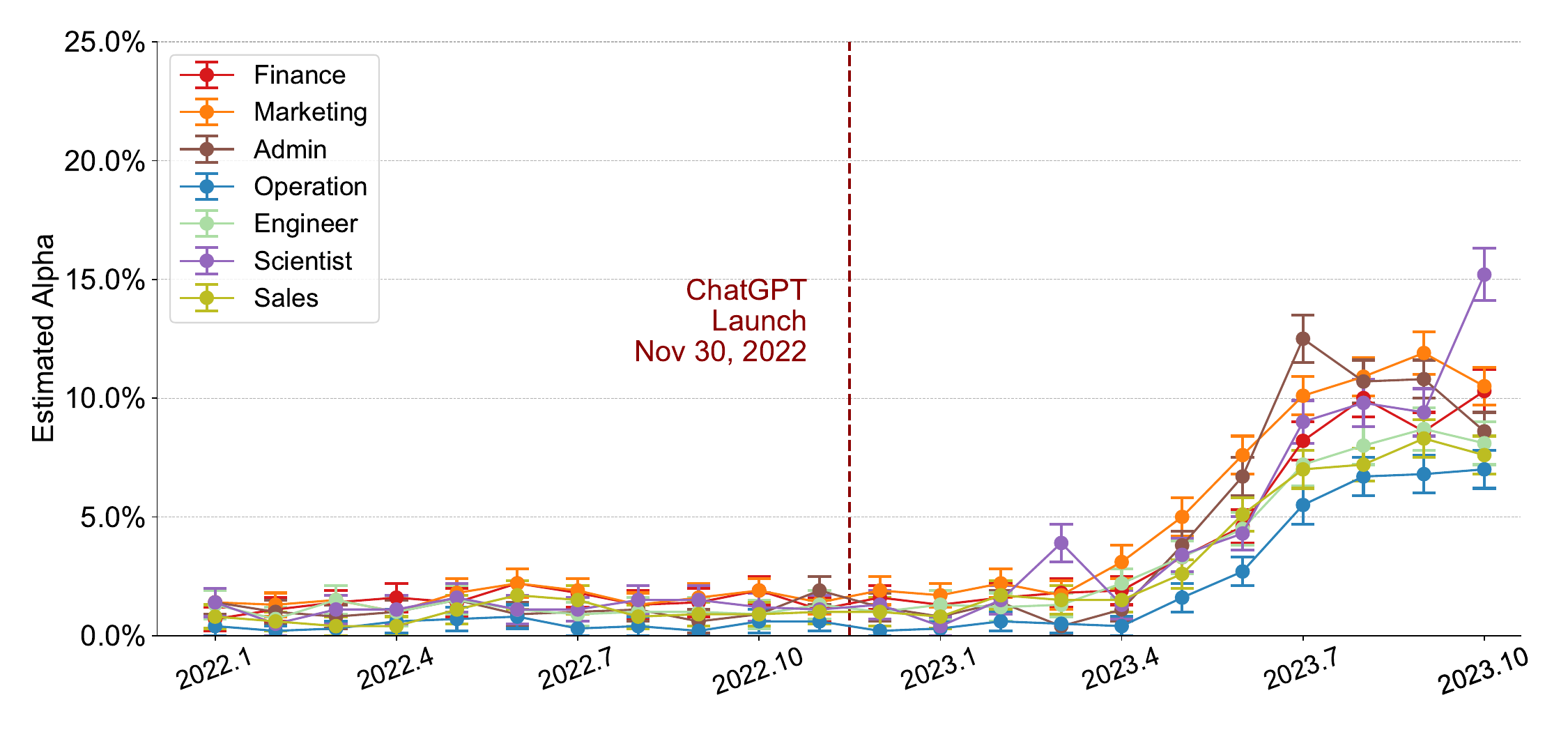}
\caption{
\textbf{LLM adoption patterns in LinkedIn job postings from small organizations ($\leq$10 employees).} 
Temporal analysis of estimated fraction ($\alpha$) of LLM-modified content across professional categories (Finance, Marketing, Admin, Operation, Engineer, Scientist, Sales) shows patterns consistent with main findings based on vacancy frequency. Following ChatGPT's launch (November 30, 2022), organizations with $\leq$10 employees demonstrate similar adoption trajectories to those posting $\leq$2 vacancies annually, with estimated $\alpha$ increasing from 0-2\% pre-launch to 7-15\% by October 2023. Scientist positions show highest adoption ($\approx$15\%), followed by Marketing and Finance ($>$10\%), while Admin, Engineer, Sales and Operations show more moderate adoption (7-9\%). Error bars indicate 95\% confidence intervals obtained through bootstrap analysis. This consistency across different definitions of small organizations (by employee count or vacancy frequency) strengthens the robustness of observed adoption patterns.
}
\label{fig: supp-robust-small-company-definition}
\end{suppfigure}

\clearpage

\begin{suppfigure}[htb]
\centering
\includegraphics[width=0.75\textwidth]{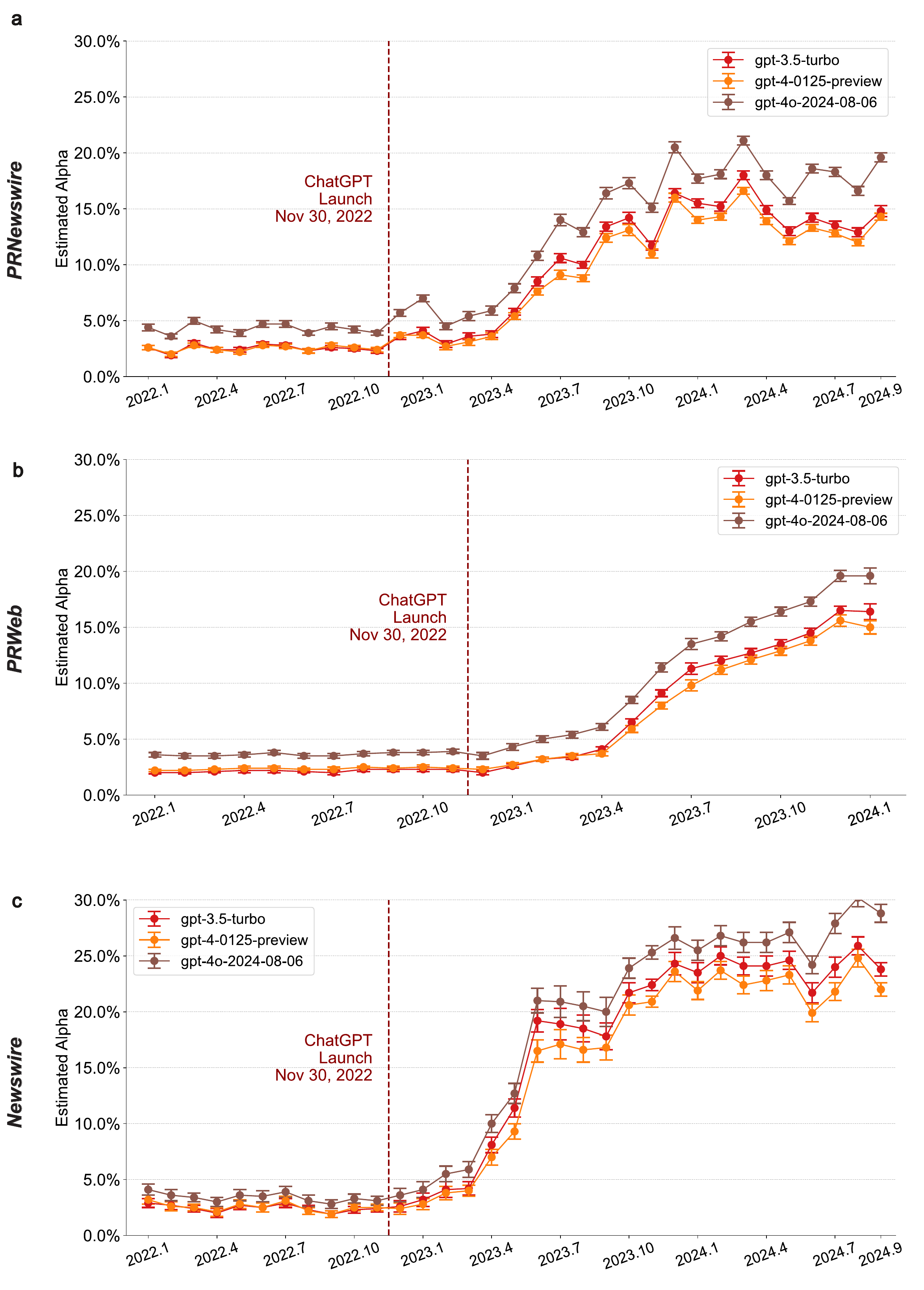}
\caption{
\textbf{Robustness analysis of LLM adoption estimates across different press release platforms using multiple GPT models for training data generation.}
(a) PRNewswire, (b) PRWeb, and (c) Newswire press releases show consistent temporal patterns regardless of the GPT model used for training data generation. Estimated fraction ($\alpha$) of LLM-modified content was calculated using three different models: GPT-3.5-turbo (used in main analysis, released January 25, 2024), GPT-4-0125-preview (released January 25, 2024), and GPT-4-2024-08-06 (released August 6, 2024). While all models reveal similar adoption trajectories following ChatGPT's launch (November 30, 2022), the most recent model GPT-4-2024-08-06 generates marginally higher estimates across platforms, suggesting our main results may be conservative. Error bars indicate 95\% confidence intervals obtained through bootstrap analysis.
}
\end{suppfigure}

\clearpage
\newpage

\begin{supptable}[htb!]
\small
\begin{center}

\caption{
\textbf{Performance validation of our model} across Consumer Complaint (all predating ChatGPT's launch), using a blend of official human and LLM-generated complaints. 
}
\label{t1}
\begin{tabular}{lrcllc}
\cmidrule[\heavyrulewidth]{1-6}
\multirow{2}{*}{\bf No.} 
& \multirow{2}{*}{\bf \begin{tabular}[c]{@{}c@{}} Validation \\ Data Source 
\end{tabular} } 
& \multirow{2}{*}{\bf \begin{tabular}[c]{@{}c@{}} Ground \\ Truth $\alpha$
\end{tabular}}  
&\multicolumn{2}{l}{\bf Estimated} 
& \multirow{2}{*}{\bf \begin{tabular}[c]{@{}c@{}} Prediction \\ Error 
\end{tabular} } 
\\
\cmidrule{4-5}
 & & & $\alpha$ & $CI$ ($\pm$) & \\
\cmidrule{1-6}
(1) & \emph{Consumer Complaint} & 0.0\% & 1.8\% & 0.2\% & 1.8\% \\
(2) & \emph{Consumer Complaint} & 2.5\% & 4.6\% & 0.2\% & 2.1\% \\
(3) & \emph{Consumer Complaint} & 5.0\% & 7.3\% & 0.2\% & 2.3\% \\
(4) & \emph{Consumer Complaint} & 7.5\% & 9.8\% & 0.2\% & 2.3\% \\
(5) & \emph{Consumer Complaint} & 10.0\% & 12.2\% & 0.3\% & 2.2\% \\
(6) & \emph{Consumer Complaint} & 12.5\% & 14.6\% & 0.2\% & 2.1\% \\
(7) & \emph{Consumer Complaint} & 15.0\% & 17.1\% & 0.3\% & 2.1\% \\
(8) & \emph{Consumer Complaint} & 17.5\% & 19.4\% & 0.3\% & 1.9\% \\
(9) & \emph{Consumer Complaint} & 20.0\% & 21.8\% & 0.3\% & 1.8\% \\
(10) & \emph{Consumer Complaint} & 22.5\% & 24.2\% & 0.3\% & 1.7\% \\
(11) & \emph{Consumer Complaint} & 25.0\% & 26.5\% & 0.3\% & 1.5\% \\
\cmidrule[\heavyrulewidth]{1-6}
\end{tabular}
\end{center}
\vspace{-5mm}
\end{supptable}

\begin{supptable}[htb!]
\small
\begin{center}

\caption{
\textbf{Performance validation of our model} across UN Press Release (all predating ChatGPT's launch), using a blend of official human and LLM-generated press releases. 
}
\label{t2}
\begin{tabular}{lrcllc}
\cmidrule[\heavyrulewidth]{1-6}
\multirow{2}{*}{\bf No.} 
& \multirow{2}{*}{\bf \begin{tabular}[c]{@{}c@{}} Validation \\ Data Source 
\end{tabular} } 
& \multirow{2}{*}{\bf \begin{tabular}[c]{@{}c@{}} Ground \\ Truth $\alpha$
\end{tabular}}  
&\multicolumn{2}{l}{\bf Estimated} 
& \multirow{2}{*}{\bf \begin{tabular}[c]{@{}c@{}} Prediction \\ Error 
\end{tabular} } 
\\
\cmidrule{4-5}
 & & & $\alpha$ & $CI$ ($\pm$) & \\
\cmidrule{1-6}
(1) & \emph{UN Press Release} & 0.0\% & 2.5\% & 0.2\% & 2.5\% \\
(2) & \emph{UN Press Release} & 2.5\% & 5.4\% & 0.2\% & 2.9\% \\
(3) & \emph{UN Press Release} & 5.0\% & 8.1\% & 0.3\% & 3.1\% \\
(4) & \emph{UN Press Release} & 7.5\% & 10.7\% & 0.3\% & 3.2\% \\
(5) & \emph{UN Press Release} & 10.0\% & 13.1\% & 0.3\% & 3.1\% \\
(6) & \emph{UN Press Release} & 12.5\% & 15.6\% & 0.3\% & 3.1\% \\
(7) & \emph{UN Press Release} & 15.0\% & 18.0\% & 0.3\% & 3.0\% \\
(8) & \emph{UN Press Release} & 17.5\% & 20.4\% & 0.3\% & 2.9\% \\
(9) & \emph{UN Press Release} & 20.0\% & 22.8\% & 0.3\% & 2.8\% \\
(10) & \emph{UN Press Release} & 22.5\% & 25.1\% & 0.3\% & 2.6\% \\
(11) & \emph{UN Press Release} & 25.0\% & 27.5\% & 0.3\% & 2.5\% \\
\cmidrule[\heavyrulewidth]{1-6}
\end{tabular}
\end{center}
\vspace{-5mm}
\end{supptable}

\begin{supptable}[htb!]
\small
\begin{center}

\caption{
\textbf{Performance validation of our model} across PRNewswire, PRWeb, Newswire (all predating ChatGPT's launch), using a blend of official human and LLM-generated press releases. 
Our algorithm demonstrates high accuracy with less than 3.3\% prediction error in identifying the proportion of LLM press release within the validation set.
}
\label{t3}
\begin{tabular}{lrcllc}
\cmidrule[\heavyrulewidth]{1-6}
\multirow{2}{*}{\bf No.} 
& \multirow{2}{*}{\bf \begin{tabular}[c]{@{}c@{}} Validation \\ Data Source 
\end{tabular} } 
& \multirow{2}{*}{\bf \begin{tabular}[c]{@{}c@{}} Ground \\ Truth $\alpha$
\end{tabular}}  
&\multicolumn{2}{l}{\bf Estimated} 
& \multirow{2}{*}{\bf \begin{tabular}[c]{@{}c@{}} Prediction \\ Error 
\end{tabular} } 
\\
\cmidrule{4-5}
 & & & $\alpha$ & $CI$ ($\pm$) & \\
\cmidrule{1-6}
(1) & \emph{PRNewswire} & 0.0\% & 2.9\% & 0.3\% & 2.9\% \\
(2) & \emph{PRNewswire} & 2.5\% & 5.7\% & 0.3\% & 3.2\% \\
(3) & \emph{PRNewswire} & 5.0\% & 8.3\% & 0.3\% & 3.3\% \\
(4) & \emph{PRNewswire} & 7.5\% & 10.8\% & 0.3\% & 3.3\% \\
(5) & \emph{PRNewswire} & 10.0\% & 13.2\% & 0.3\% & 3.2\% \\
(6) & \emph{PRNewswire} & 12.5\% & 15.6\% & 0.3\% & 3.1\% \\
(7) & \emph{PRNewswire} & 15.0\% & 18.0\% & 0.3\% & 3.0\% \\
(8) & \emph{PRNewswire} & 17.5\% & 20.3\% & 0.3\% & 2.8\% \\
(9) & \emph{PRNewswire} & 20.0\% & 22.7\% & 0.3\% & 2.7\% \\
(10) & \emph{PRNewswire}& 22.5\% & 25.0\% & 0.3\% & 2.5\% \\
(11) & \emph{PRNewswire}& 25.0\% & 27.3\% & 0.3\% & 2.3\% \\
\cmidrule{1-6}
(12) & \emph{PRWeb} & 0.0\% & 2.1\% & 0.2\% & 2.1\% \\
(13) & \emph{PRWeb} & 2.5\% & 5.2\% & 0.2\% & 2.7\% \\
(14) & \emph{PRWeb} & 5.0\% & 7.8\% & 0.2\% & 2.8\% \\
(15) & \emph{PRWeb} & 7.5\% & 10.4\% & 0.2\% & 2.9\% \\
(16) & \emph{PRWeb} & 10.0\% & 12.9\% & 0.3\% & 2.9\% \\
(17) & \emph{PRWeb} & 12.5\% & 15.4\% & 0.3\% & 2.9\% \\
(18) & \emph{PRWeb} & 15.0\% & 17.8\% & 0.3\% & 2.8\% \\
(19) & \emph{PRWeb} & 17.5\% & 20.2\% & 0.3\% & 2.7\% \\
(20) & \emph{PRWeb} & 20.0\% & 22.6\% & 0.3\% & 2.6\% \\
(21) & \emph{PRWeb} & 22.5\% & 25.0\% & 0.3\% & 2.5\% \\
(22) & \emph{PRWeb} & 25.0\% & 27.3\% & 0.3\% & 2.3\% \\
\cmidrule{1-6}
(23) & \emph{Newswire} & 0.0\% & 2.3\% & 0.2\% & 2.3\% \\
(24) & \emph{Newswire} & 2.5\% & 5.3\% & 0.2\% & 2.8\% \\
(25) & \emph{Newswire} & 5.0\% & 7.9\% & 0.3\% & 2.9\% \\
(26) & \emph{Newswire} & 7.5\% & 10.5\% & 0.3\% & 3.0\% \\
(27) & \emph{Newswire} & 10.0\% &13.0\% & 0.3\% & 3.0\% \\
(28) & \emph{Newswire} & 12.5\% & 15.4\% & 0.3\% & 2.9\% \\
(29) & \emph{Newswire} & 15.0\% & 17.9\% & 0.3\% & 2.9\% \\
(30) & \emph{Newswire} & 17.5\% & 20.3\% & 0.3\% & 2.8\% \\
(31) & \emph{Newswire} & 20.0\% & 22.6\% & 0.3\% & 2.6\% \\
(32) & \emph{Newswire} & 22.5\% & 25.0\% & 0.3\% & 2.5\% \\
(33) & \emph{Newswire} & 25.0\% & 27.4\% & 0.3\% & 2.4\% \\
\cmidrule[\heavyrulewidth]{1-6}
\end{tabular}
\end{center}
\vspace{-5mm}
\end{supptable}

\begin{supptable}[htb!]
\small
\begin{center}

\caption{
\textbf{Performance validation of our model} across Admin, Engineer, Finance, Marketing (all predating ChatGPT's launch), using a blend of official human and LLM-generated job postings. 
}
\label{t4}
\begin{tabular}{lrcllc}
\cmidrule[\heavyrulewidth]{1-6}
\multirow{2}{*}{\bf No.} 
& \multirow{2}{*}{\bf \begin{tabular}[c]{@{}c@{}} Validation \\ Data Category 
\end{tabular} } 
& \multirow{2}{*}{\bf \begin{tabular}[c]{@{}c@{}} Ground \\ Truth $\alpha$
\end{tabular}}  
&\multicolumn{2}{l}{\bf Estimated} 
& \multirow{2}{*}{\bf \begin{tabular}[c]{@{}c@{}} Prediction \\ Error 
\end{tabular} } 
\\
\cmidrule{4-5}
 & & & $\alpha$ & $CI$ ($\pm$) & \\
\cmidrule{1-6}
(1) & \emph{Admin} & 0.0\% & 1.2\% & 0.5\% & 1.2\% \\
(2) & \emph{Admin} & 2.5\% & 4.0\% & 0.6\% & 1.5\% \\
(3) & \emph{Admin} & 5.0\% & 6.6\% & 0.7\% & 1.6\% \\
(4) & \emph{Admin} & 7.5\% & 9.1\% & 0.7\% & 1.6\% \\
(5) & \emph{Admin} & 10.0\% & 11.6\% & 0.8\% & 1.6\% \\
(6) & \emph{Admin} & 12.5\% & 14.1\% & 0.8\% & 1.6\% \\
(7) & \emph{Admin} & 15.0\% & 16.7\% & 0.8\% & 1.7\% \\
(8) & \emph{Admin} & 17.5\% & 19.1\% & 0.8\% & 1.6\% \\
(9) & \emph{Admin} & 20.0\% & 21.6\% & 0.9\% & 1.6\% \\
(10) & \emph{Admin}& 22.5\% & 24.0\% & 0.9\% & 1.5\% \\
(11) & \emph{Admin}& 25.0\% & 26.4\% & 0.9\% & 1.4\% \\
\cmidrule{1-6}
(12) & \emph{Engineer} & 0.0\% & 0.9\% & 0.5\% & 0.9\% \\
(13) & \emph{Engineer} & 2.5\% & 3.6\% & 0.6\% & 1.1\% \\
(14) & \emph{Engineer} & 5.0\% & 6.2\% & 0.7\% & 1.2\% \\
(15) & \emph{Engineer} & 7.5\% & 8.8\% & 0.8\% & 1.3\% \\
(16) & \emph{Engineer} & 10.0\% & 11.3\% & 0.8\% & 1.3\% \\
(17) & \emph{Engineer} & 12.5\% & 13.8\% & 0.8\% & 1.3\% \\
(18) & \emph{Engineer} & 15.0\% & 16.4\% & 0.9\% & 1.4\% \\
(19) & \emph{Engineer} & 17.5\% & 18.9\% & 0.8\% & 1.4\% \\
(20) & \emph{Engineer} & 20.0\% & 21.4\% & 0.9\% & 1.4\% \\
(21) & \emph{Engineer} & 22.5\% & 23.9\% & 0.9\% & 1.4\% \\
(22) & \emph{Engineer} & 25.0\% & 26.4\% & 0.9\% & 1.4\% \\
\cmidrule{1-6}
(23) & \emph{Finance} & 0.0\% & 0.7\% & 0.4\% & 0.7\% \\
(24) & \emph{Finance} & 2.5\% & 3.5\% & 0.6\% & 1.0\% \\
(25) & \emph{Finance} & 5.0\% & 6.0\% & 0.7\% & 1.0\% \\
(26) & \emph{Finance} & 7.5\% & 8.5\% & 0.7\% & 1.0\% \\
(27) & \emph{Finance} & 10.0\% &10.9\% & 0.7\% & 0.9\% \\
(28) & \emph{Finance} & 12.5\% & 13.4\% & 0.7\% & 0.9\% \\
(29) & \emph{Finance} & 15.0\% & 15.9\% & 0.8\% & 0.9\% \\
(30) & \emph{Finance} & 17.5\% & 18.3\% & 0.8\% & 0.8\% \\
(31) & \emph{Finance} & 20.0\% & 20.7\% & 0.9\% & 0.7\% \\
(32) & \emph{Finance} & 22.5\% & 23.1\% & 0.8\% & 0.6\% \\
(33) & \emph{Finance} & 25.0\% & 25.5\% & 0.9\% & 0.5\% \\
\cmidrule{1-6}
(23) & \emph{Marketing} & 0.0\% & 0.6\% & 0.5\% & 0.6\% \\
(24) & \emph{Marketing} & 2.5\% & 3.4\% & 0.6\% & 0.9\% \\
(25) & \emph{Marketing} & 5.0\% & 5.9\% & 0.6\% & 0.9\% \\
(26) & \emph{Marketing} & 7.5\% & 8.4\% & 0.7\% & 0.9\% \\
(27) & \emph{Marketing} & 10.0\% &10.9\% & 0.8\% & 0.9\% \\
(28) & \emph{Marketing} & 12.5\% & 13.4\% & 0.8\% & 0.9\% \\
(29) & \emph{Marketing} & 15.0\% & 15.8\% & 0.8\% & 0.8\% \\
(30) & \emph{Marketing} & 17.5\% & 18.3\% & 0.9\% & 0.8\% \\
(31) & \emph{Marketing} & 20.0\% & 20.8\% & 0.8\% & 0.8\% \\
(32) & \emph{Marketing} & 22.5\% & 23.3\% & 0.9\% & 0.8\% \\
(33) & \emph{Marketing} & 25.0\% & 25.7\% & 0.9\% & 0.7\% \\
\cmidrule[\heavyrulewidth]{1-6}
\end{tabular}
\end{center}
\vspace{-5mm}
\end{supptable}
\clearpage
\newpage
\begin{supptable}[htb!]
\small
\begin{center}

\caption{
\textbf{Performance validation of our model} across Operation, Sales, Scientist (all predating ChatGPT's launch), using a blend of official human and LLM-generated job postings. 
}
\label{t5}
\begin{tabular}{lrcllc}
\cmidrule[\heavyrulewidth]{1-6}
\multirow{2}{*}{\bf No.} 
& \multirow{2}{*}{\bf \begin{tabular}[c]{@{}c@{}} Validation \\ Data Category 
\end{tabular} } 
& \multirow{2}{*}{\bf \begin{tabular}[c]{@{}c@{}} Ground \\ Truth $\alpha$
\end{tabular}}  
&\multicolumn{2}{l}{\bf Estimated} 
& \multirow{2}{*}{\bf \begin{tabular}[c]{@{}c@{}} Prediction \\ Error 
\end{tabular} } 
\\
\cmidrule{4-5}
 & & & $\alpha$ & $CI$ ($\pm$) & \\
\cmidrule{1-6}
(1) & \emph{Operation} & 0.0\% & 0.8\% & 0.5\% & 0.8\% \\
(2) & \emph{Operation} & 2.5\% & 3.3\% & 0.6\% & 0.8\% \\
(3) & \emph{Operation} & 5.0\% & 5.9\% & 0.7\% & 0.9\% \\
(4) & \emph{Operation} & 7.5\% & 8.4\% & 0.7\% & 0.9\% \\
(5) & \emph{Operation} & 10.0\% & 10.9\% & 0.8\% & 0.9\% \\
(6) & \emph{Operation} & 12.5\% & 13.3\% & 0.8\% & 0.8\% \\
(7) & \emph{Operation} & 15.0\% & 15.8\% & 0.8\% & 0.8\% \\
(8) & \emph{Operation} & 17.5\% & 18.2\% & 0.9\% & 0.7\% \\
(9) & \emph{Operation} & 20.0\% & 20.7\% & 0.9\% & 0.7\% \\
(10) & \emph{Operation}& 22.5\% & 23.2\% & 0.9\% & 0.7\% \\
(11) & \emph{Operation}& 25.0\% & 25.6\% & 0.9\% & 0.6\% \\
\cmidrule{1-6}
(12) & \emph{Sales} & 0.0\% & 1.2\% & 0.5\% & 1.2\% \\
(13) & \emph{Sales} & 2.5\% & 3.7\% & 0.6\% & 1.2\% \\
(14) & \emph{Sales} & 5.0\% & 6.2\% & 0.7\% & 1.2\% \\
(15) & \emph{Sales} & 7.5\% & 8.6\% & 0.8\% & 1.1\% \\
(16) & \emph{Sales} & 10.0\% & 11.0\% & 0.8\% & 1.0\% \\
(17) & \emph{Sales} & 12.5\% & 13.4\% & 0.8\% & 0.9\% \\
(18) & \emph{Sales} & 15.0\% & 15.8\% & 0.8\% & 0.8\% \\
(19) & \emph{Sales} & 17.5\% & 18.2\% & 0.8\% & 0.7\% \\
(20) & \emph{Sales} & 20.0\% & 20.7\% & 0.9\% & 0.7\% \\
(21) & \emph{Sales} & 22.5\% & 23.1\% & 0.9\% & 0.6\% \\
(22) & \emph{Sales} & 25.0\% & 25.5\% & 0.9\% & 0.5\% \\
\cmidrule{1-6}
(23) & \emph{Scientist} & 0.0\% &  2.0\% & 0.6\% & 2.0\% \\
(24) & \emph{Scientist} & 2.5\% &  4.8\% & 0.7\% & 2.3\% \\
(25) & \emph{Scientist} & 5.0\% &  7.3\% & 0.7\% & 2.3\% \\
(26) & \emph{Scientist} & 7.5\% &  9.8\% & 0.8\% & 2.3\% \\
(27) & \emph{Scientist} & 10.0\% & 12.3\% & 0.8\% & 2.3\% \\
(28) & \emph{Scientist} & 12.5\% & 14.7\% & 0.9\% & 2.2\% \\
(29) & \emph{Scientist} & 15.0\% & 17.2\% & 0.9\% & 2.2\% \\
(30) & \emph{Scientist} & 17.5\% & 19.7\% & 1.0\% & 2.2\% \\
(31) & \emph{Scientist} & 20.0\% & 22.1\% & 0.9\% & 2.1\% \\
(32) & \emph{Scientist} & 22.5\% & 24.5\% & 1.0\% & 2.0\% \\
(33) & \emph{Scientist} & 25.0\% & 27.0\% & 1.0\% & 2.0\% \\
\cmidrule[\heavyrulewidth]{1-6}
\end{tabular}
\end{center}
\vspace{-5mm}
\end{supptable}

\newpage
\clearpage

\newcounter{mysuppfigure}
\newenvironment{mysuppfigure}[1][]{%
  \addtocounter{mysuppfigure}{1}%
  \renewcommand{\thesuppfigure}{S\arabic{mysuppfigure}}%
  \begin{suppfigure}[#1]%
}{%
  \end{suppfigure}%
}

\newcounter{mysupptable}
\newenvironment{mysupptable}[1][]{%
  \addtocounter{mysupptable}{1}%
  \renewcommand{\thesupptable}{S\arabic{mysupptable}}%
  \begin{supptable}[#1]%
}{%
  \end{supptable}%
}

\section*{Supplementary Information}

\subsection*{Overview of the Consumer Complaint Data}
\label{main:subsec:Consumer Complaint-data}

The Consumer Complaint Database, maintained by the Consumer Financial Protection Bureau (CFPB), is a publicly accessible resource that collects complaints about consumer financial products and services. These complaints are forwarded to companies for their response, while the CFPB—a U.S. government agency—is dedicated to ensuring that banks, lenders, and other financial institutions treat consumers fairly. We focus on 687,241 consumer complaint narrative, starting from January 2022 and ending in August 2024. The dataset offers the mailing ZIP code provided by the consumer, which allow us to check heterogeneity via the educational level and the degree of urbanization by region. Specifically, we employ Rural Urban Commuting Area (RUCA) codes to assess urbanization levels and measure the educational level by the percentage of individuals aged 25 and older who have earned a bachelor’s degree. Corresponding data is available at 
\href{https://www.ers.usda.gov/data-products/rural-urban-commuting-area-codes}{\texttt{here}}
and \href{https://data.census.gov/table/ACSST1Y2023.S1501}{\texttt{here}} respectively.

\subsection*{Overview of the LinkedIn Job Posting Data}
\label{main:subsec:job-posting-data}

We use data from the Revelio Labs universe, which collects, cleans and aggregates individual-level job postings sourced from publicly available online sources, such as LinkedIn. The raw dataset includes all LinkedIn postings (active, inactive, removed), the company identifier, the company founding year, the full text of job listings, and associated information (title, salary, etc.).  The raw data are broken out by Revelio Labs into eight job categories: Administration, Engineering, Finance, Marketing, Operations, Sales, Scientist, and Unclassified. We focus on 304,270,122 job postings, starting from January 2021 and ending in October 2023. We focus on the full text of the job postings. To analyze the heterogeneity of LLM usage by company characteristics, we combine the job listings information with the Revelio Labs associated LinkedIn employee data. Similarly to the job postings data, the baseline workforce data was scraped, cleaned and aggregated at the firm level. The workforce data is available going back up to 2008. We define firm characteristics based on pre-ChatGPT introduction characteristics. We define two different definitions for small firms: in our sample, small firms are companies with either 10 or fewer registered employees in 2021 or companies posting less than or equal to about 2 postings per year. We also check heterogeneity via founding year, splitting in terms of years 2015-onwards, 2000-2015, 1980-2000 and before 1980. These time periods are determined based on quantiles of the founding year distribution. Note that although the median number of postings per company per year is 3, the total number of postings drops from 304,270,122 to 1,440,912 when we focus on small companies. This indicates that small companies contribute a relatively minor share to the total posting volume compared to larger companies.

\subsection*{Overview of the Corporate Press Release Data}
\label{main:subsec:press-release-data}

We collect corporate press release data using the NewsAPI service, which aggregates online news content from various sources. We collected data from: PRNewswire, PRWeb, and Newswire, three of the main companies distributing corporate press releases online. These were chosen due to data avilability and cost. PR Newswire, founded in 1954, is one of the oldest and most widely recognized press release distribution services, offering an extensive network that reaches major news outlets, journalists, and online platforms worldwide. It serves a broad range of clients, from large corporations to small businesses. PRWeb, launched in 1997, focuses primarily on online distribution and SEO optimization, making it a more budget-friendly option for businesses looking to enhance their digital presence. Newswire distributes press releases to both traditional media and online platforms, catering to businesses of various sizes. While all three services offer some level of editorial support, their primary business focus remains distribution.

With a focus on English-language text, we gathered up to 537,413 press releases from January 2022 to September 2024.  Our analysis primarily focused on the full body text. Due to the limited number of articles post-ChatGPT introduction available from Newswire, we conducted detailed robustness checks only on PR Newswire and PRWeb data, which provided sufficient volume for heterogeneity analysis.  We  classified the press releases by four overarching categories: Business \& Money, Science \& Tech, People \& Culture, and Other.

\subsection*{Overview of the UN Press Release Data}
\label{main:subsec:press-release-data}

We collect United Nations release data using customized scripts. The United Nations (UN), founded in 1945, is an international organization dedicated to fostering global peace, security, and cooperation among its member states~\cite{shin2024adoption}. 
Country teams of United Nations regularly update on the latest developments in that country.
To ensure consistency and maintain a focus on English-language content, articles were selected from the English-language websites of 97 country teams. From January 2019 to September 2024, up to 15,919 press releases were collected, with the analysis primarily concentrating on the full body text. Our investigation revealed that among the remaining 96 country teams, 57 do not have their own websites, 33 lack English-language websites, and 6 do not operate press release websites.

\subsection*{Data Split, Model Fitting, and Evaluation}
\label{main:subsec:training-validation}

For model fitting, we count word frequencies for the corpora written before the release of ChatGPT and the LLM-modified corpora. We fit the model with data from 2021 (2019 for UN press release), and use data from January 2022 onwards for validation and inference. We developed individual models for each major category in LinkedIn job postings and for each distribution platform in corporate press releases. For UN press releases and consumer complaints, we fit one model for each domain. During inference, we randomly sample up to 2,000 records per month (per quarter for UN press release) to analyze the increasing temporal trends of LLM usage across various writing domains.

To evaluate model accuracy and calibration under temporal distribution shift, we collected a sample of 2000 records from January 1, 2022, to November 29, 2022, a time period prior to the release of ChatGPT, as the validation data. We construct validation sets with LLM-modified content proportions ($\alpha$) ranging from 0\% to 25\%, in 2.5\% increments, and compared the model's estimated $\alpha$ with the ground truth $\alpha$ (Table \ref{t1}, \ref{t2}, \ref{t3}, \ref{t4}, \ref{t5}). Our models all performed well in our application, with a prediction error consistently less than 3.3\% at the population level across various ground truth $\alpha$ values.

\clearpage
\newpage

\begin{suppfigure}[htb!]
\begin{lstlisting}
The aim here is to reverse-engineer the author's writing process by taking a piece of text from a consumer complaint and compressing it into a more concise form. This process simulates how an author might distill their thoughts and key points into a structured, yet not overly condensed form. 

Now as a first step, given a complete piece of text from a consumer complaint, reverse-engineer it into a list of bullet points.
\end{lstlisting}
\caption{
Example prompt for summarizing a consumer complaint into a skeleton: This process simulates how an author might first only write the main ideas and core information into a concise outline. The goal is to capture the essence of the complaint in a structured and succinct manner, serving as a foundation for the next prompt.
}
\label{fig:skeleton-prompt-1}
\end{suppfigure}

\begin{suppfigure}[htb!]
\begin{lstlisting}
Following the initial step of reverse-engineering the author's writing process by compressing a text segment from a consumer complaint, you now enter the second phase. Here, your objective is to expand upon the concise version previously crafted. This stage simulates how an author elaborates on the distilled thoughts and key points, enriching them into a detailed, structured narrative. 

Given the concise output from the previous step, your task is to develop it into a fully fleshed-out text.
\end{lstlisting}
\caption{
Example prompt for expanding the skeleton into a full text: The aim here is to simulate the process of using the structured outline as a basis to generate comprehensive and coherent text. This step mirrors the way an author might flesh out the outline into detailed paragraphs, effectively transforming the condensed ideas into a fully articulated consumer complaint. The format and depth of the expansion can vary, reflecting the diverse styles and requirements of different consumer complaints.
}
\label{fig:skeleton-prompt-2}
\end{suppfigure}

\end{document}